\pdfoutput=1
\documentclass[letterpaper, 10 pt, conference]{ieeeconf}  

\IEEEoverridecommandlockouts                              

\usepackage{amsthm}
\usepackage{graphicx}
\usepackage{caption}
\usepackage{subcaption}
\usepackage{caption}
\usepackage[linesnumbered,ruled,vlined]{algorithm2e}
\usepackage{algorithmicx}
\usepackage{amsmath}
\theoremstyle{plain}
\usepackage{mathtools}
\usepackage{epsfig}
\usepackage{amssymb}
\usepackage{epstopdf}
\usepackage{multicol}
\usepackage{multirow}
\usepackage{array}
\usepackage{url}
\usepackage{color}
\usepackage{cite}
\usepackage{float}
\usepackage{wrapfig}
\usepackage{hyperref}

\title{\LARGE \bf
Dynamically Constrained Motion Planning Networks \\for Non-Holonomic Robots
}

\author{Jacob J. Johnson, Linjun Li, Fei Liu, Ahmed H. Qureshi and Michael C. Yip
\thanks{J. J. Johnson, L. Li, F. Liu, A. H. Qureshi and M. C. Yip are with Department of Electrical and Computer Engineering, University of California San Diego, La Jolla, CA 92093 USA. {\tt\small \{jjj025,lili,f4liu,a1qureshi,yip\}@ucsd.edu}}%
}

\begin{document}

\maketitle
\thispagestyle{empty}
\pagestyle{empty}

\begin{abstract}
Reliable real-time planning for robots is essential in today's rapidly expanding automated ecosystem. In such environments, traditional methods that plan by relaxing constraints become unreliable or slow-down for kinematically constrained robots. This paper describes the algorithm Dynamic Motion Planning Networks (Dynamic MPNet), an extension to Motion Planning Networks, for non-holonomic robots that address the challenge of real-time motion planning using a neural planning approach. We propose modifications to the training and planning networks that make it possible for real-time planning while improving the data efficiency of training and trained models' generalizability. We evaluate our model in simulation for planning tasks for a non-holonomic robot. We also demonstrate experimental results for an indoor navigation task using a Dubins car.

\end{abstract}
\section{Introduction}
The problem of motion planning is among the core robotics challenges, which aims to find a path between given start and goal states while satisfying a set of desired constraints \cite{lavalle1998rapidly}. In most cases, the desired constraints to be met are merely collision-avoidance. However, with the increasing familiarity and adoption of mobile robots and autonomous vehicles, constrained kinematics such as non-holonomic constraints have become prevalent in many situations that require the planner not just to perform collision-avoidance but also adhere to the system's dynamical equations \cite{laumond1998guidelines}. 

Non-holonomic constraints are governed by dynamical equations which depend on the time-derivative of the system's configuration space \cite{laumond1998guidelines}. These constraints arise in many applications, ranging from mobile robot navigation \cite{laumond1998guidelines} to needle steering in robot surgery \cite{alterovitz2007stochastic}. To generalize, it comes in cases where the system's control space is of a lower-dimension than its configuration space. For instance, in car-like robots, the control inputs are the linear and angular velocities, while the configuration/robot-motion space is three-dimensional (x/y position and heading). Consequently, a feasible trajectory in the robot's configuration space might not be feasible with respect to the system's dynamics \cite{laumond1998guidelines, de1998feedback}.

\begin{figure}[th!]
    \centering
    \includegraphics[width=\columnwidth]{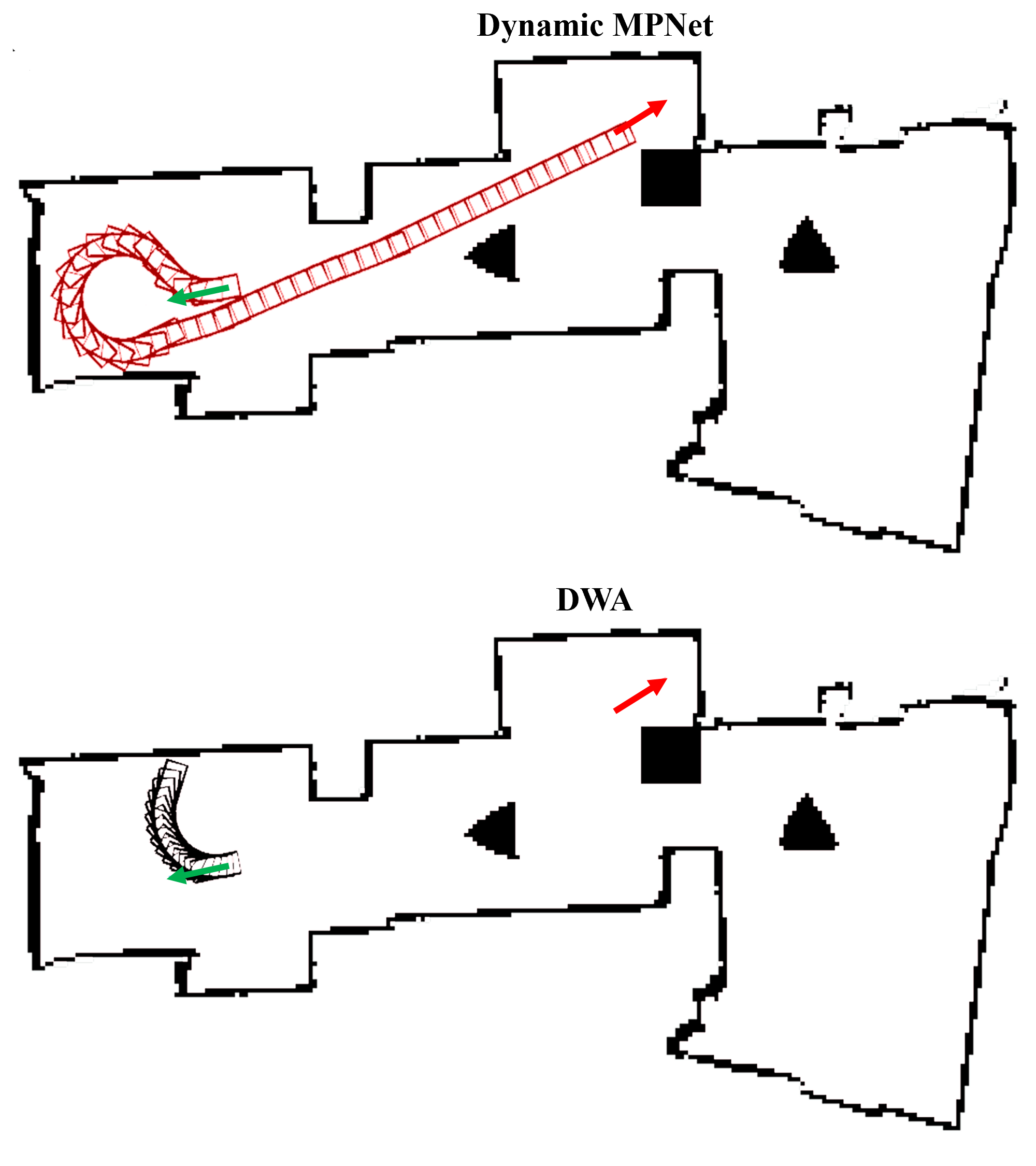}
    \caption{The trajectory the robot followed for a given start (green arrow) and goal (red arrow) position using, Dynamic MPNet (red), and Dynamic Windows Approach (black) as local planners respectively. Since the DWA planner has no kinematic constraints, it is difficult for the planner to generate paths with U-turns, while since the path generated from Dynamic MPNet encodes kinematic constraints, it is able to generate a successful path towards the goal.}
    \label{fig:Uturn}
\end{figure}

Various algorithms have been proposed to address real-time motion planning under non-holonomic constraints,  on the spectrum of Sampling-based Motion Planning (SMP) to non-Sampling-based methods\cite{xu2012real, rosmann2017kinodynamic, karaman2013sampling}. The advantage of sampling-based planners is that they provide probabilistic completeness, i.e., the probability that the planner finds a solution reaches one as the number of samples reaches infinity. Sampling-based planners were used for real-time planning for non-holonomic systems in the form of anytime planners\cite{karaman2011anytime}, aiming to improve a current plan while executing an already evaluated plan\cite{karaman2011anytime}. Although such planners can easily find a feasible path in uncluttered spaces, they often fail to find a solution in reasonable time in constrained spaces as well as need to relax goal constraints (e.g., final orientation).

Recently reinforcement learning based methods that use neural networks have gained traction in solving motion planning problems \cite{zhang2017deep, faust2018prm, chiang2018learning} for non-holonomic systems. These methods often require careful fine-tuning of reward functions as well as a significant computing resources to search for proper hyperparameters. Another class of motion planning algorithms, called neural motion planners, have emerged that learns to imitate an oracle planner and exhibits virtues of an ideal planner during online execution \cite{qureshi2019motion, qureshi2019motionb, qureshi2018deeply}. Motion Planning Networks (MPNet) \cite{qureshi2019motion} is one of the first and prominent neural motion planning methods, showing orders of magnitude performance computational speed compared to previous offline methods while producing near-optimal solutions. However, MPNet, along with its extensions \cite{jurgenson2019harnessing}, only considers collision-avoidance constraints and finds a viable path in the robot's configuration spaces, i.e., without taking into account the system's kinematic limitations. Furthermore, these methods also  require significant training data to achieve generalizability and, in their current formulation, have yet to be scaled to real-time planning scenarios involving navigating large environments. Thus none of these methods single-handedly has features of an ideal planner, i.e., find near-optimal/optimal paths with high, almost real-time, computational speed and exhibit completeness guarantees.

In this paper, we propose Dynamic Motion Planning Networks (Dynamic MPNet), which extends MPNet to plan under a broad class of non-holonomic constraints in real-time. Dynamic MPNet is a deep neural-network-based iterative planning algorithm. It takes the sub-goals between given start and goal states from a global C-space planner and finds a kinematically-feasible path between them with a high-computational speed and completeness guarantees. Real-time planning is made possible by planning during executing the previous plan, similar to anytime planning systems. We evaluate our framework on Dubins car dynamical model in challenging navigation tasks where state-of-the-art classical methods are failing, including both simulation and real-robot experiments.
Our results show that Dynamic MPNet outperforms existing methods in terms of accuracy and average speed for each trajectory, and, similar to MPNet, also generalizes to new planning problems. We also release Dynamic MPNet as an open-source package in the ROS navigation stack for computationally-efficient local path planning under various kinematic constraints\footnote{\href{https://github.com/jacobjj/mpnet_local_planner}{https://github.com/jacobjj/mpnet\_local\_planner}}.

\section{Problem Definition}

The motion planning algorithm with dynamic constraints can be described as follows. Consider a dynamic system defined by the differential equation $\dot{s}(t) = f(s(t), u(t))$, where $s(t)\in\mathcal{S}\subset \mathbb{R}^n$ describes the state of the system, $u(t)\in\mathcal{U}\subset \mathbb{R}^m$ describes the control inputs to the system and $\dot{s}(t)\in\mathbb{R}^n$ describes the rate of change of the system state. The state space $\mathcal{S}$ can be split into two regions: $\mathcal{S}_{obs}$ states with obstacles and $\mathcal{S}_{free}=\mathcal{S}-\mathcal{S}_{obs}$ obstacle free regions. The goal of the planning algorithm is to find a set of control inputs $u(t)$ that would move the robot from a given initial state $s_{start}\in\mathcal{S}_{free}$ to a given goal state $s_{goal}\in\mathcal{S}_{free}$, such that the sequence of actions $u(t)$ and $s(t)$  $\forall t\in[0,\tau]$ satisfy the given dynamic equation and the states are in free space, i.e. $s(t)\in \mathcal{S}_{free}$. The problem can be expanded further by adding optimality criteria to the sequence of states. In the subsequent sections, we describe our solution to the motion planning problem. The planner is used as a local planner in a hierarchical navigation architecture. 

\section{Methods}
\begin{figure}
    \centering
    \includegraphics[width=\columnwidth]{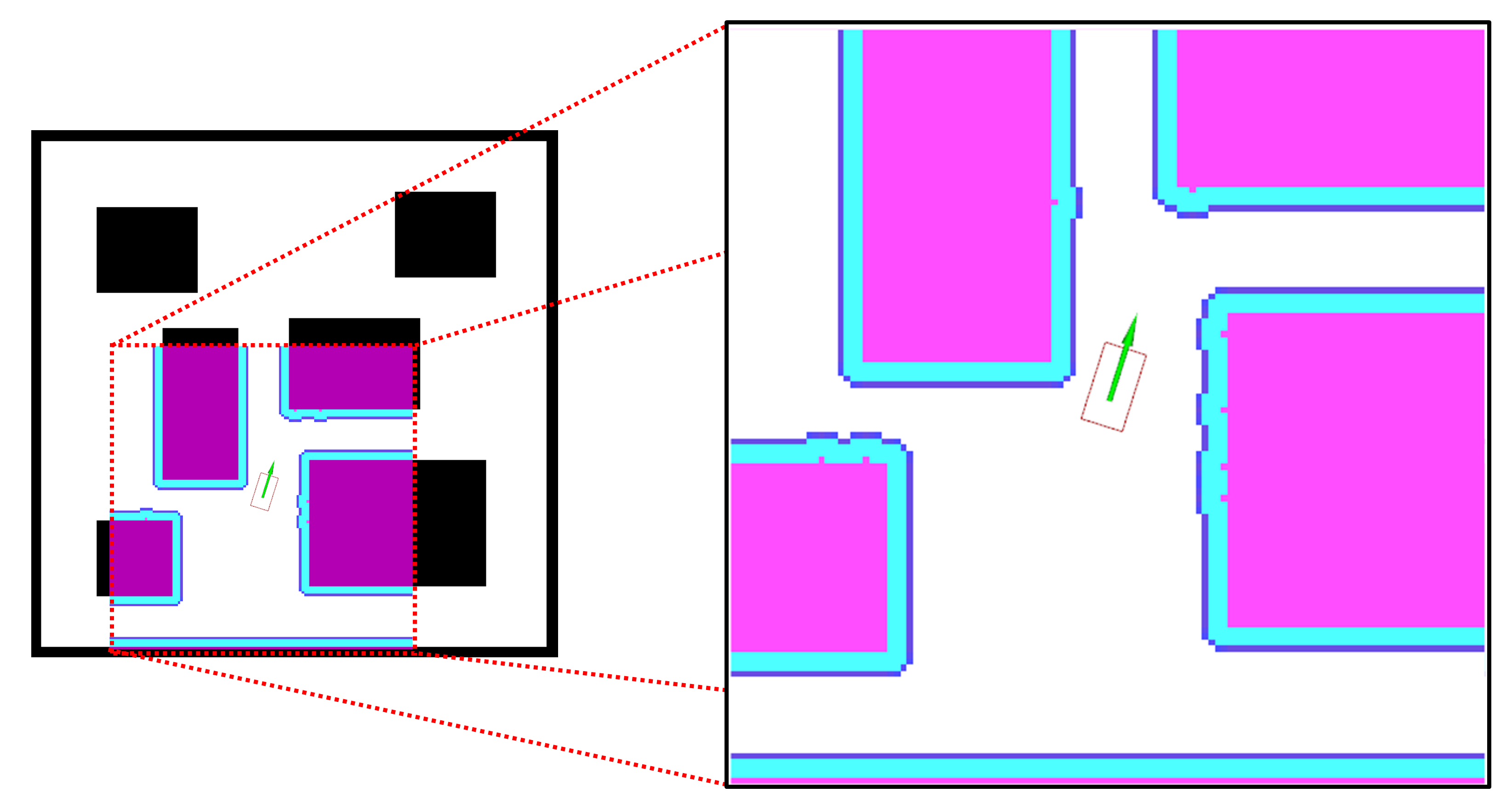}
    \caption{The area of the costmap passed to MPNet for planning. The local costmap used for planning is always egocentric to the robot. The blue region indicates the obstacles inflated by the robot footprint.}
    \label{fig:localcostmap}
\end{figure}

\begin{figure*}
    \centering
    \includegraphics[width=2\columnwidth]{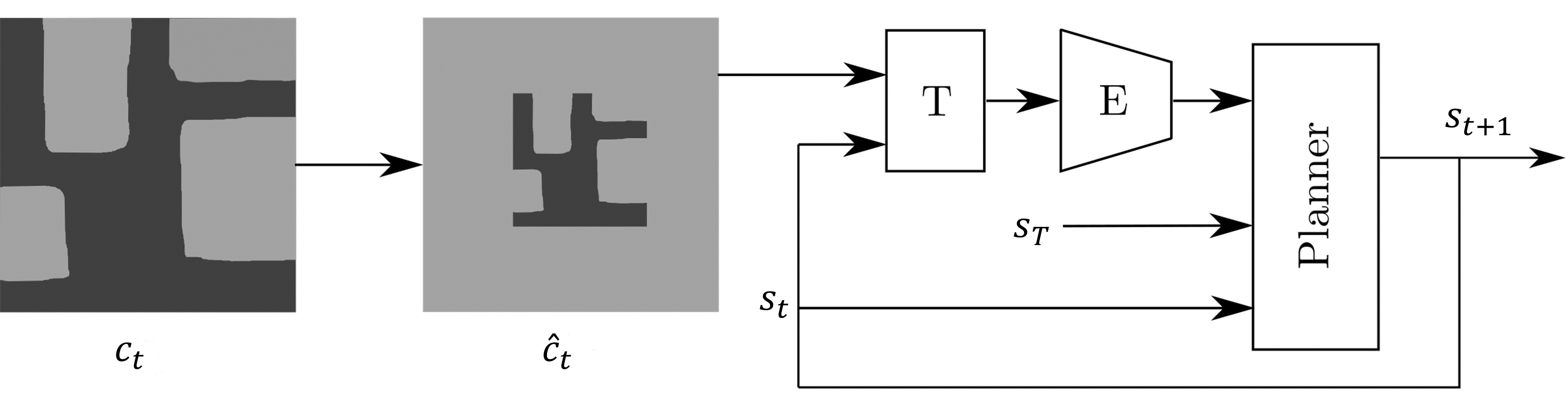}
    \caption{The graph describes the flow of inputs and outputs for the planner. $x_t$, $x_{t+1}$ and $x_g$ represents the current, next and target positions respectively. $x_g$ is the sub-goal position from the global plan. $C_t$ and, $\hat{C}_t$ is the costmap before and after padding respectively. The $T$ block centers the padded costmap, $\hat{C}_t$, with respect to the robot position $x_t$. $E$ block consists of convolution networks that encode the costmap into latent space vectors. The latent space representation of the costmap, the current robot and goal position are passed to the Planner node to generate the new target point.}
    \label{fig:network_architecture}
\end{figure*}

Dynamic MPNet uses supervised learning to train neural networks that generate near-optimal paths through an environment given a start and goal position. The networks are  trained with expert trajectories in randomized, diverse environments so that unseen environments can be planned in with near-expert level cost and with substantially less sampling than classical sample-based planners. 

Dynamic MPNet consists of 3 core modules, a \textit{transformer} that centers the local costmap with respect to the robot, an \textit{encoder} network that takes the ego-centered obstacle map and converts it into a latent vector, and a \textit{planner} network that takes the latent encoding, the current or predicted pose, and goal pose and returns the next feasible step. In the following section, we will go over the environment encoding, network training and planning pipeline in more detail.
\subsection{Environment Encoding}
Most non-holonomic systems reside in environments that can vary significantly in size. Encoding an environment where the map could be arbitrarily large in size is infeasible. This is where a hierarchical approach to the navigation is useful. Using the global plan as a guide, kinematically feasible paths are generated for a local region using Dynamic MPNet. Fig. \ref{fig:localcostmap} is an example for the local costmap passed into the network planner. This costmap is always egocentric to the robot (see Fig. \ref{fig:localcostmap}).  Doing this reduces the dimensionality of the input space, as the output of the network is only a function of the current robot orientation, relative goal position, and costmap. It could also be viewed as a normalization step, to disentangle the dependence of training trajectory and world map with the sampled point. It thus helps the model to generalize to the environment better with fewer training samples. 

\subsection{Training}
The planner is trained with dynamically feasible trajectories such that, during prediction, it will tend to predict dynamically feasible intermediate poses. Given an expert trajectory $\{s_0, s_1 ,\ldots, s_T\}$ that satisfies the dynamical constraints of the robot for the planning problem, the network uses the current position ($s_t$), goal position ($s_T$) and obstacle representation ($\hat{c_t}$) to predict the next state ($\hat{s}_{t+1}$). These four elements form a training tuple ($s_t, s_T, \hat{c_t}, s_{t+1}$). The encoding and planning networks are trained by reducing the mean-squared error between the predicted state $\hat{s}_{t+1}$ and the actual state from the expert planner $s_{t+1}$ in an end-to-end fashion using gradient descent. The loss function for $N$ such trajectories is given by:

\begin{equation}
    L (\theta) = \frac{1}{N\ T} \sum_{j=1}^{N}\sum_{t=1}^{T-1}||s_{j,t} - \hat{s}_{j,t}||^2
\end{equation}
where $\theta$ represents the combined parameters of both the encoder and planner network.


\subsection{Planning}
The network starts planning using the current position, sub-goal position from the global plan, and the local costmap to generate a sequence of kinematically feasible states. For each step, if the predicted state is kinematically feasible and is collision-free, it is used as the current position for the next prediction. Fig. \ref{fig:network_architecture} shows the complete pipeline. Then, for each sampled step during the planning, an egocentric costmap encoding at this immediate (non-initial) location is required to generate the next step. This costmap is achieved by translating the initial costmap. It is necessary to pad the obstacle map such that no loss in obstacle information occurs after transforming the map to an egocentric pose of an intermediate step of a motion plan. Given a grid of size $l\times l$, we pad it to make a new grid of size $2l\times 2l$. This way, for the planner, the world's perception remains the same, since all the padded spaces are still obstacles. 


Algorithm \ref{mpnetalgorithm} outlines the Dynamic MPNet planner, and Algorithm \ref{neuralplanner} outlines the modified path generation heuristic. The functionality of the different function calls are defined as follows:

\subsubsection{Padding} The \verb|Pad| function takes a costmap $c_{obs}\in \mathbb{R}^{l\times l}$, and returns a padded costmap $\hat{c}_{temp}\in \mathbb{R}^{2l\times 2l}$. Padded values are assumed to be obstacle regions to prevent the planner to find paths that would navigate this non-physical space. See Fig. \ref{fig:network_architecture} for an example.

\subsubsection{Steering} The function \verb|Steer| ($x_1,x_2$) checks if we can generate a sequence of kinematically feasible states without collision from $x_1$ to $x_2$ within a fixed time. It returns a feasible path if it exists otherwise, an empty list. For non-holonomic systems, a differential equation solver or parameterized curves along with a collision checker can be used to implement this function.

\subsubsection{Network} The function \verb|Net| represents both the encoder and planner neural network combined. It generates the next possible point on the path, given the current and goal position of the robot and the modified costmap. The flow of inputs is described in Fig. \ref{fig:network_architecture}.

\subsubsection{Add} The function \verb|Add| ($\tau, x_1$) appends the path $\tau$ with the node $x_1$.

\subsubsection{Transform} Given a point $x_i$ and a padded costmap $\hat{c}$, the function \verb|Transform| ($\hat{c}$, $x_i$) translates the costmap in such a way that the costmap is egocentric to the position $x_i$.

\subsubsection{Replanning} Only if, for a given start and goal location, the neural planner is not able to provide a feasible path under a suitable amount of time, the function will run classical sampling-based methods until a path is found, specifically to ensure probabilistic completeness.
\begin{algorithm}
\SetAlgoLined
\SetKwFunction{Pad}{Pad}
\SetKwFunction{NeuralPlanner}{NeuralPlanner}
\SetKwFunction{Replanner}{Replanner}
\SetKwFunction{Empty}{Empty}
    $\hat{c}\gets$ \Pad{$c_{obs}$}\;
    $\tau\gets$ \NeuralPlanner{$x_{start}, x_{goal},\hat{c}$}\;
    \If{\Empty{$\tau$}}{
        $\tau\gets$\Replanner{$x_{start}, x_{goal}$}\;
    }
    \Return{$\tau$}
\caption{$\tau\gets$DynamicMPNet ($x_{start}$, $x_{goal}$, $c_{obs}$)}
\label{mpnetalgorithm}
\end{algorithm}

\begin{algorithm}
\SetAlgoLined
\SetKwFunction{Steer}{Steer}
\SetKwFunction{Net}{Net}
\SetKwFunction{Add}{Add}
\SetKwFunction{Transform}{Transform}
\SetKwFunction{Notempty}{NotEmpty}
    $\tau\gets \{x_{from}\}$\;
    \For{$i=0$ \KwTo $N$}{
    $x_{temp} \gets$ \Net{$x_{from}, x_{goal}, \hat{c}$}\;
        $\tau_{temp}$ $\gets$ \Steer{$x_{from}, x_{temp}$}\;
        \If{\Notempty{$\tau_{temp}$}}{
            $\tau\gets$\Add{$\tau,\tau_{temp}$}\;
            $\tau_{goal}$ $\gets $\Steer{$x_{temp},x_{goal}$}\;
            \If{\Notempty{$\tau_{goal}$}}{
                $\tau \gets$\Add{$\tau, \tau_{temp}$}\;
                \Return{$\tau$}
            }
            $x_{from}\gets x_{temp}$\;
            $\hat{c}\gets$\Transform{$\hat{c},x_{from}$}\;
        }
    }
    \Return{$\varnothing$}
\caption{$\tau\gets$NeuralPlanner ($x_{from}$, $x_{to}$, $\hat{c}$)}
\label{neuralplanner}
\end{algorithm}

\section{Experiments}
In this section, we describe the data collection, model architecture and experiment setup used in this paper. The neural networks model was defined and trained using the PyTorch\cite{pytorch} python library, and the trained model was loaded into C++ using the torch C++ API. To test the viability of our framework, we integrated the Dynamic MPNet planner to the navigation stack\cite{navigation_stack} provided by the ROS community and compared it with standard local planners used by the community. We used the default global planner from the ROS navigation stack and the mit-racecar\cite{mit-racecar} model was used as the robot for the simulations.\footnote{\href{https://youtu.be/1b3i1SSiUms}{https://youtu.be/1b3i1SSiUms}}

\subsection{Data Collection}
\begin{figure*}
    \centering
    \includegraphics[width=2\columnwidth]{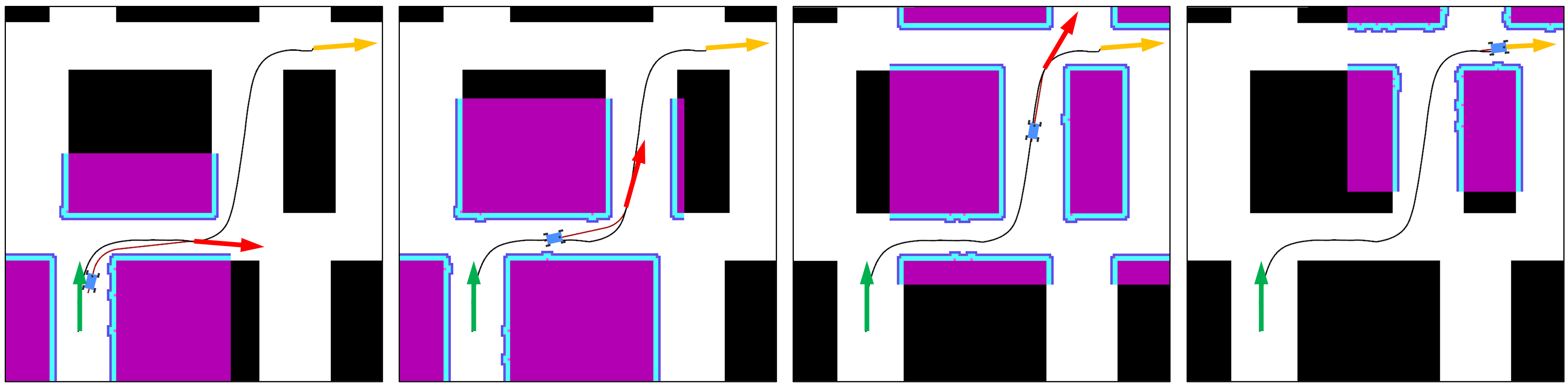}
    \caption{For a given start (green arrow) and goal (orange arrow) position, the plan generated by the Dynamic MPNet (red path) for a given sub-goal (red arrow). The black trajectory is the global plan. The colored region represents the local costmap used by the Dynamic MPNet.}
    \label{fig:gridworldPath}
\end{figure*}

For training the model, we collected expert trajectories using the RRT*\cite{lavalle1998rapidly} algorithm using the Open Source Motion Planning Library\cite{sucan2012the-open-motion-planning-library}. For the grid map (Fig. \ref{fig:gridworldPath}) environment, we trained the model on 10,000 RRT* trajectories by randomly sampling a start and goal location for a fixed time, while for the real-world environment, we generated 12,000 trajectories. To further augment our data set, we chose smaller sections of a path and added it to the training data. Thus if we have $n$ points on a path, then we can choose any 2 points and create $\mathcal{O} (\frac{n^2}{4})$ trajectories. The data collection was accelerated by encapsulating the sampling code in a docker container and launching multiple containers concurrently with different seed values.
\subsection{Model Architecture}
A convolutional neural network model was used for environment encoding. Prior works in robotics have also used CNN's to process the cost map for planning \cite{planningEgoHolger} and path prediction \cite{Baumann2018PredictingEP}.  The input to the encoder was an $l\times l$ dimension cost map. The CNN consisted of 3 convolutional layers, with kernel size $[5,5]$, $[3,3]$, and $[3,3]$, and output channels of 8,16 and 32, respectively. A maxpool and Parametric Rectified Linear Unit\cite{Prelu} (PReLU) layer follow the first two convolutional layers. The output of the final convolutional layer is passed through a PReLU layer to generate the output of the encoder.

The planner was a fully connected neural network with six layers. A PReLU\cite{Prelu} and a Dropout\cite{dropout} layer follow the first four hidden layers. Dropout is used not only during training to prevent overfitting\cite{dropout} but also during prediction to introduce stochasticity that tends to make motion planning networks more robust\cite{qureshi2019motion}. A PReLU follows the penultimate hidden layer and the output of the final hidden layer is passed through a tanh nonlinearity. Both networks were trained in an end-to-end fashion.

\subsection{Kinematic Model of a Car-Like Robot and Dubins curve}
In this paper we consider non-holonomic kinematics following the Dubins vehicle  model\cite{Dubins1957OnCO} though in practice the constraint can be of another variety. The Dubins model is given by: 
\begin{equation}
    \mathbf{\dot{s}(t)} = 
        \begin{bmatrix}
        \dot{x}(t) \\ \dot{y}(t) \\ \dot{\theta}(t)
        \end{bmatrix} = 
        \begin{bmatrix}
        v_s\cos(\phi) \\ v_s\sin(\phi) \\ \frac{v_s}{d}\tan{\phi} 
        \end{bmatrix}
    \label{dubinsCar}
\end{equation}
where $v_s$ is the speed of the car, $\phi$ is the steering angle and $d$ the distance between the rear and front axle. 
We use Dubins curves as a steering function for Dynamic MPNet. Given the state variables for a Dubins vehicle, the shortest path is a unique path among 6-basis  trajectories\cite{Dubins1957OnCO}. Each of these trajectories is represented by a sequence of left, right and straight turns. To steer between given two states, the shortest among the 6-trajectories are used. 
\subsection{Trajectory Tracking}
Given a sparsely sampled path produced by Dynamic MPNet, a trajectory tracking problem is then presented during runtime to move the vehicle between the sampled points. This is necessary to ensure that due to unmodelled effects and under noise and disturbances that the vehicle follows, to the best of its capability, the solution to a dynamically feasible path provided by Dynamic MPNet.

We formulate the trajectory tracking problem as a constrained discrete optimization problem with a finite horizon. In order to follow the trajectory generated from MPNet, we sample the path nodes adaptively to proper sparsity with similar distances and apply a Nonlinear Model Predictive Control (NMPC, \cite{NMPC}) with the following objective function:
\begin{equation}
    \begin{aligned}
    \text{minimize} &\sum_{t=0}^{N}{w_s\left\lVert \mathbf{s(t)} - \mathbf{\hat{s}(t)}\right\lVert  + w_u\left\lVert \phi(t)\right\lVert) + w_a\left\lVert \Delta\phi(t)\right\lVert}  \\ 
    \text{subject to} &\ \Delta \mathbf{s(t)} = \begin{bmatrix} 
                                    v_s\cos(\phi) \\ v_s\sin(\phi) \\ \frac{v_s}{d}\tan{\phi}
                                \end{bmatrix} \Delta t
    \end{aligned}
\end{equation}
where $\Delta t$ and $\Delta s$ are time step and state difference respectively, $w_s$ is state loss weight, $w_u$, $w_a$ are weights for control loss, $\hat{s}$ are sampled path nodes from Dynamic MPNet, $v_s$ is preset as constant velocity and $N$ is the prediction horizon. In each control cycle, the control output is computed by solving the problem with Interior Point Optimizer (Ipopt, \cite{ipopt,Wachter2006}) implemented with algorithmic differentiation library CppAD \cite{10.1007/978-3-540-68942-3_7} and the first action is taken from solution. 

The prediction horizon was determined using path curvature and maximum velocity and was in sync with the control frequency to reach the targeted state. The $w_a$ term contributes to decreasing the vibration due to maneuvering, which guarantees the smoothness of the trajectory.

\section{Results}
We evaluate the learned model in simulation on seen and unseen environments, and a real-world indoor environment using a Dubins car robot.The Dubins car is set up similar to the MIT Racecar\cite{mit-racecar}. The local planners available for car-like robots in the ROS-navigation stack is  the Dynamic-Window Approach (DWA) planner\cite{fox1997dynamic} and Timed-Elastic-Band (TEB)\cite{TEBcar}. Although the TEB is the only ROS local planner for car-like robots, the optimization problem was not able to generate paths for Dubins car. As a result we compared our algorithm with DWA. We also implemented an Anytime RRT* local planner by generating the path from RRT* rather than from the Neural Planner. We compare the robot's accuracy and average speed over a batch of planning problems. Each planning problem was verified to have a solution using an offline RRT* planner. For each test case, the problem is considered solved if the robot center is able to achieve the target position within a radius of 0.2m (about half the length of the robot) and target orientation within 15$^\circ$. The same threshold is used for all local planners. Each planner runs at 5Hz, giving it 200 ms to find and optimize a feasible plan. In the following sections we report our results from our experiments.

To evaluate the sampling speed of Dynamic MPNet on the trained simulated environment in terms of execution time and path length, we measured the time taken to sample $n$ points using Dynamic MPNet is given in Table \ref{tab:worstCase}. These times indicate that the Dynamic MPNet planner is able to generate a path within 20Hz if set to high resolution of 50 points per local path generated. Thus Dynamic MPNet is able to generate kinematically reachable points in real-time. 
\begin{table}[th]
    \centering
    \captionsetup{justification=centering}
    \caption{Planning Time of Dynamic MPNet vs. Sampling Resolution}
    \begin{tabular}{c|cccc}
        \hline
         Number of samples & 5 & 10 & 25 & 50  \\
         Compute Time (ms) & 11.33 & 15.29 & 22.07 & 44.67  \\
         \hline
    \end{tabular}
    \label{tab:worstCase}
\end{table}
\begin{figure}
    \centering
    \includegraphics[width=0.9\columnwidth]{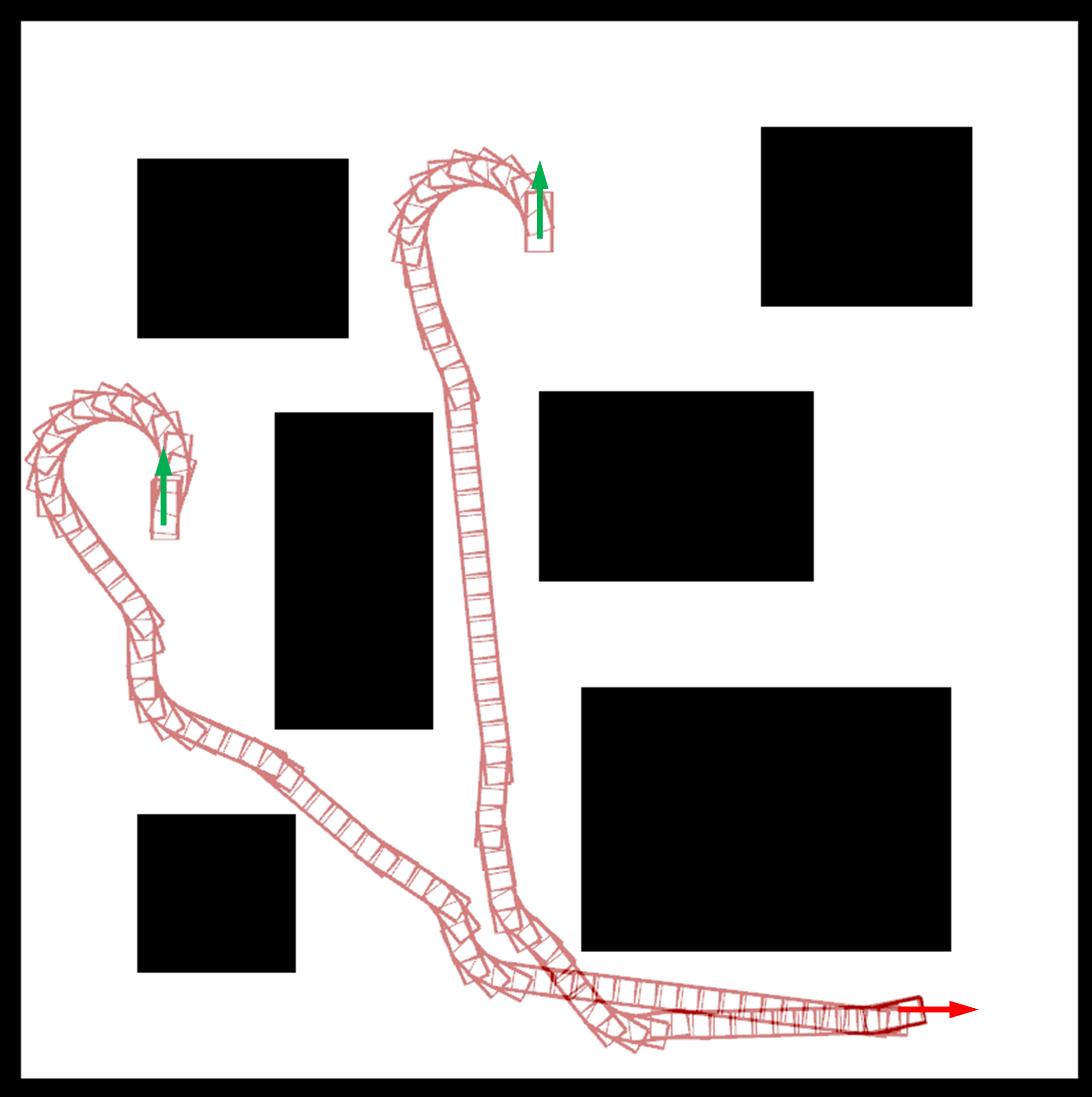}
    \caption{The Dynamic MPNet generating trajectories for an untrained map for two different start and end goal on a synthetic map. The planner can generate trajectories that comply with the kinematic constraints of the robot, thus achieving higher accuracy compared to DWA.}
    \label{fig:Dynamic MPNetUnknown}
    \vspace{-6mm}
\end{figure}

The Dynamic MPNet was trained on a synthetic grid world environment. One of the paths generated by the trained planner is shown in Fig. \ref{fig:gridworldPath}. To evaluate the generalizability of the planner, we created a synthetic map that is different from the training map, but shares a lot of common obstacle features such as $90^\circ$ turns from the original map. Paths generated on this environment is shown in Fig. \ref{fig:Dynamic MPNetUnknown}. Table \ref{tab:gridWorld} compares the percentage of planning problems solved with standard planners. Dynamic MPNet is able to plan 34\% more planning problems compared to DWA on the unseen map. Hence we were able to achieve generalizabilty with much fewer training paths.


\begin{table}
    \captionsetup{justification=centering}
    \caption{Percentage of Planning Problems Successfully Completed}
    \centering
    \begin{tabular}{cccc}
        \hline
        Environment & DWA & Anytime RRT* &  Dynamic MPNet\\
        \hline 
        Unseen map & 47\% & 74\% & 80\% \\
        Real world map & 44\% & 58\% & 76\%\\
        \hline
    \end{tabular}
    \label{tab:gridWorld}
\end{table}

\begin{figure}
    \centering
    \includegraphics[width=0.9\columnwidth]{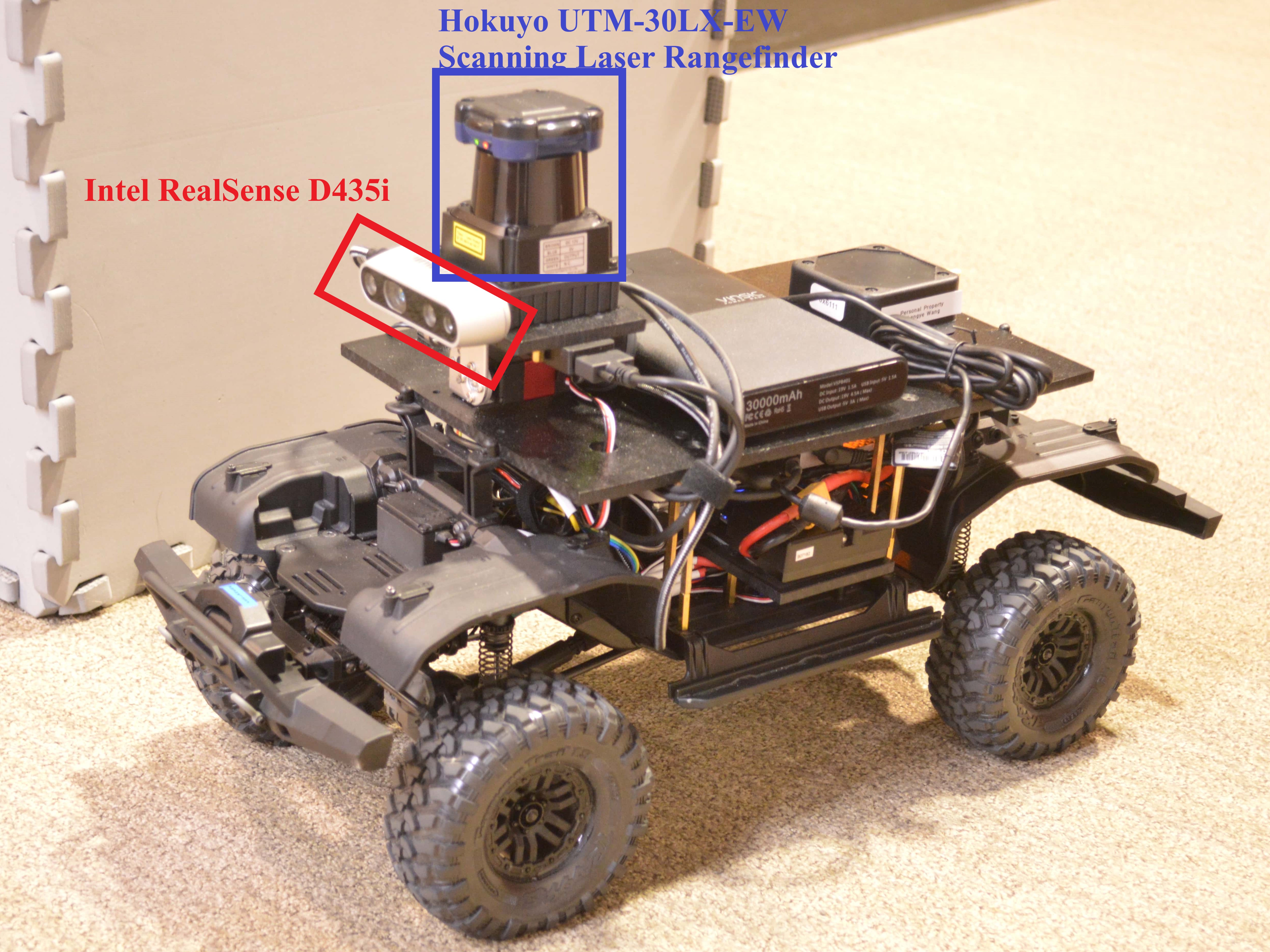}
    \caption{The RC robot used for this work. On board LIDAR (Hokuyo UTM-30LX-EW) and IMU (RealSense D435i) sensors were used for localization.}
    \label{fig:rcrobot}
    \vspace{-6mm}
\end{figure}

\begin{figure}
    \centering
    \includegraphics[width=\columnwidth]{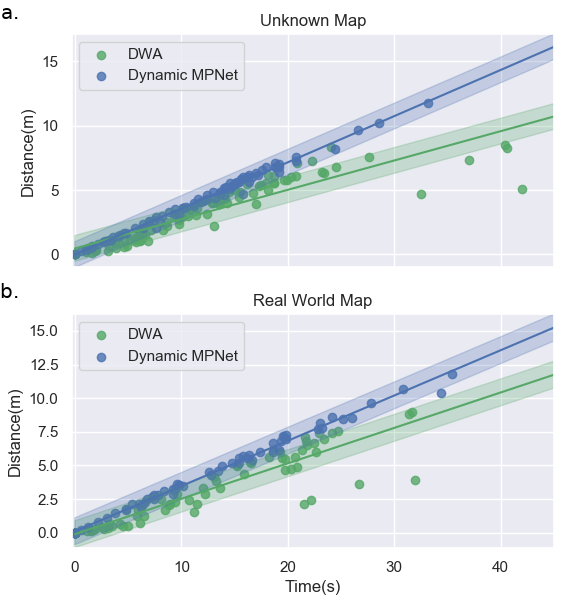}
    \caption{Total time and total distance taken by Dynamic MPNet and DWA planners on the same set of planning problems for an (a) unknown map (b) real world map. A linear regression model was fit to each dataset to evaluate the average speed of the car for the local planners. The shaded region represents standard error in the estimates. Dynamic MPnet not only moves faster and more consistently than DWA, it solves problems that require longer planning distances as well.}
    \label{fig:slope_comp}
    \vspace{-2mm}
\end{figure}

\begin{figure}
    \centering
    \includegraphics[width=0.95\columnwidth]{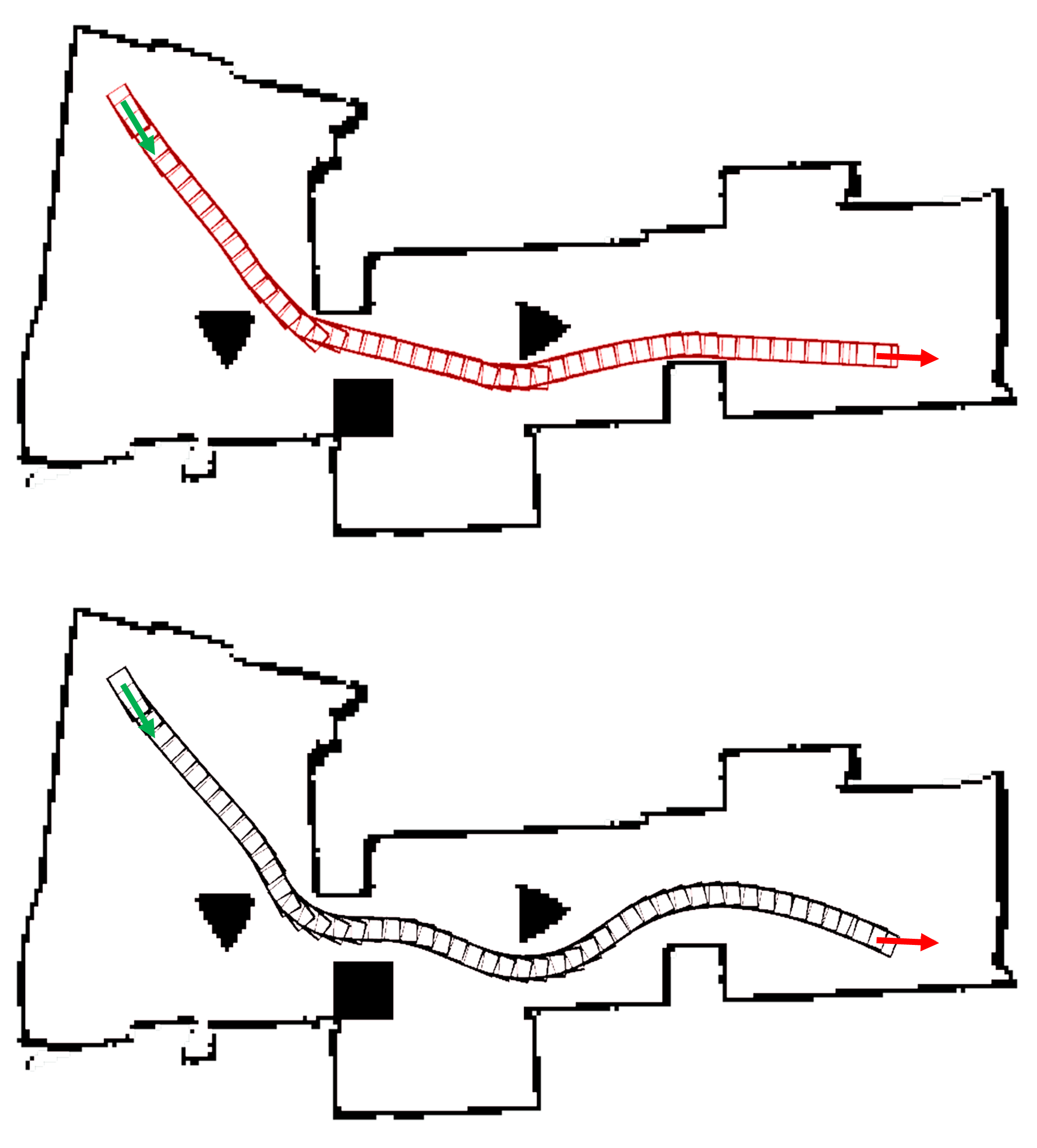}
    \caption{The trajectory the robot followed for a given start (green arrow) and goal (red arrow) position for DWA (black) and Dynamic MPNet (red). The path Dynamic MPNet takes is much closer to the obstacle compared to DWA and hence shorter.}
    \label{fig:mymapPath}
    \vspace{-6mm}
\end{figure}

In addition to the simulation experiments, real-world experiments with an RC Dubins Car (see Fig. \ref{fig:rcrobot}) in one of our mapped office buildings were used. Fig. \ref{fig:mymapPath} compares one of the trajectories planned by DWA and Dynamic MPNet for the same start and goal point. The red path given by Dynamic MPNet is 9.43m long while the DWA path is 10.85m long. Since Dynamic MPNet is trained on RRT* paths, the local paths generated would be near optimal. In Fig. \ref{fig:slope_comp} we compare the distance of each trajectory and time take to complete the planning problems solved by both DWA and Dynamic MPNet for the unknown and real world maps. A linear regression model was fit to both the models to estimate the average speed of the robot. Table \ref{tab:avgspeed} summarizes the results. Dynamic MPNet is able to solve planning problems faster compared to DWA.
\begin{table}
    \centering
    \captionsetup{justification=centering}
    \caption{Average Vehicle Speed of Dynamic MPnet v.s. DWA}
    \begin{tabular}{ccc}
        \hline
        Environment & DWA Speed (m/s) & Dynamic MPNet Speed (m/s)\\
        \hline 
        Unseen map & 0.211 $\pm$ 0.016 & 0.340 $\pm$ 0.006 \\
        Real world map &  0.263 $\pm$ 0.014 & 0.336 $\pm$ 0.003\\
        \hline
    \end{tabular}
    \label{tab:avgspeed}
\end{table}

In Fig. \ref{fig:Uturn} we can observe one of the biggest drawbacks of the DWA planner which is that it does not take the kinematics constraint of the robot into consideration for generating a plan. Since the Dynamic MPNet generates kinodynamically feasible paths, it is able to plan for the given goal position and orientation. As a result, MPNet is able to solve a larger percentage of planning problems compared to standard planners for both simulated and real world maps.

\section{Conclusions}
In this work, we have used neural motion planners to generate paths for non-holonomic robots successfully. We achieved this by introducing a  new framework to facilitate real-time planning for non-holonomic robots. Compared to traditional sampling-based methods, Dynamic MPNet can achieve  faster average speeds with higher accuracy. 
In our future studies, one of our primary goals is to extend Dynamic MPNet to problems with moving obstacles by leveraging its remarkable properties of finding near-optimal paths in almost real-time computation speed. One of the short coming of real time planning  method is it's dependence on the global plan to generate a feasible path to the goal because of which real time navigation methods fail on a large number of planning problems for non-holonomic systems. A future goal would be to integrate the constraint planner to generate a kinematically feasible global plan. Another future objective is to utilize Dynamic MPNet for needle steering in surgical tasks that, in most cases, require a planner to satisfy underlying non-holonomic constraints.

\bibliographystyle{IEEEtran}
\bibliography{root}
\nocite{*}
\end{document}